\documentclass[10pt,twocolumn,letterpaper]{article}

\usepackage[review]{cvpr}      

\usepackage{graphicx}
\newcommand{\partitle}[1]{\smallskip \noindent \textbf{#1.}}
\usepackage{microtype}
\usepackage{enumitem}
\usepackage{amsmath,amsbsy,amsfonts,amssymb,amsthm,units,bm,mathrsfs}

\usepackage{algorithm, algorithmic}
\usepackage{booktabs}
\usepackage{multirow}
\usepackage{array}
\usepackage{amsfonts}
\usepackage{subfiles}

\theoremstyle{definition}

\newtheorem{definition}{Definition}[section]

\newtheorem{remark}{Remark}

\makeatletter
\newsavebox\myboxA
\newsavebox\myboxB
\newlength\mylenA

\newcommand*\xbar[2][0.75]{%
    \sbox{\myboxA}{$\m@th#2$}%
    \setbox\myboxB\null
    \ht\myboxB=\ht\myboxA%
    \dp\myboxB=\dp\myboxA%
    \wd\myboxB=#1\wd\myboxA
    \sbox\myboxB{$\m@th\overline{\copy\myboxB}$}
    \setlength\mylenA{\the\wd\myboxA}
    \addtolength\mylenA{-\the\wd\myboxB}%
    \ifdim\wd\myboxB<\wd\myboxA%
       \rlap{\hskip 0.5\mylenA\usebox\myboxB}{\usebox\myboxA}%
    \else
        \hskip -0.5\mylenA\rlap{\usebox\myboxA}{\hskip 0.5\mylenA\usebox\myboxB}%
    \fi}
\makeatother

\newcommand{\vect}[1]{\ensuremath{\mathbf{#1}}}
\newcommand{\mat}[1]{\ensuremath{\mathbf{#1}}}

\newcommand{\argmax}{\mathop{\rm argmax}}

\newcommand{\expect}{\mathbb{E}}

\newcommand{\V}{\mat{V}}

\newcommand{\Q}{\mat{Q}}
\newcommand{\E}{\mat{E}}

\newcommand{\x}{\vect{x}}

\usepackage[pagebackref,breaklinks,colorlinks]{hyperref}

\usepackage[capitalize]{cleveref}
\crefname{section}{Sec.}{Secs.}
\Crefname{section}{Section}{Sections}
\Crefname{table}{Table}{Tables}
\crefname{table}{Tab.}{Tabs.}

\begin{document}

\title{SNEAK: Synonymous Sentences-Aware Adversarial Attack on Natural Language Video Localization}

\maketitle

\begin{abstract}

Natural language video localization (NLVL) is an important task in the vision-language understanding area, which calls for an in-depth understanding of not only computer vision and natural language side alone, but more importantly the interplay between both sides. Adversarial vulnerability has been well-recognized as a critical security issue of deep neural network models, which requires prudent investigation. Despite its extensive yet separated studies in video and language tasks, current understanding of the adversarial robustness in vision-language joint tasks like NLVL is less developed. This paper therefore aims to comprehensively investigate the adversarial robustness of NLVL models by examining three facets of vulnerabilities from both attack and defense aspects. To achieve the attack goal, we propose a new adversarial attack paradigm called synonymous sentences-aware adversarial attack on NLVL (SNEAK), which captures the cross-modality interplay between the vision and language sides.
To further enhance the stealthiness of SNEAK, we propose a frame importance-guided pruning mechanism to reduce the amount of perturbed frames. Extensive experiments on two NLVL models and two datasets demonstrate the effectiveness of the proposed attacks and defense. Our implementation can be accessed via an anonymous link\footnote{\url{https://github.com/shiwen1997/SNEAK-CAF2}} and will be made publicly available.
\end{abstract}

\section{Introduction}
\label{sec:intro}
Natural language video localization (NLVL) as a typical task in the vision-language understanding area has gained increasing research interest in recent years driven by various applications in practice, e.g., crime detection\cite{c:crime}, video surveillance\cite{c:surveillance}, and vehicle retrieval\cite{c:vehicle}. NLVL models seek to predict the start and end moment in an untrimmed video that semantically corresponds to a given natural language query. Apparently, for such cross-modal vision-language tasks, a comprehensive characterization from all three following aspects is indispensable for NLVL models to achieve sound and satisfactory performance: 1) the video in the vision aspect; 2) the text query in the natural language aspect; 3) and more importantly the aspect of the cross-modal interaction between the video and text query. 

Adversarial vulnerability is among the key characteristics of models (especially Deep Neural Networks (DNN) models) that demand prudent examination, since otherwise it can raise serious security concerns for models in security-sensitive application scenarios \cite{c:505, c:506, c:507}.
For either vision or natural language areas alone, there exists an extensive research on their adversarial robustness from both attack and defense perspectives. For the vision side, most works focus on models dealing with image datasets\cite{DBLP:journals/corr/abs-2111-04266, DBLP:journals/corr/abs-2109-15009}, while in recent years increasing research efforts have also been made for the video\cite{DBLP:journals/corr/abs-2111-05468, DBLP:journals/corr/abs-2110-01823}, 3D point cloud, etc.. For the natural language side, existing works reveal that NLP models are vulnerable when facing word substitution, character substitution\cite{DBLP:conf/acl/QiYXLS20}, etc..  However, it is much less explored when it comes to the vision-language joint areas. The only existing efforts limit to image-text data-related models, where the adversarial vulnerability of dense captioning and visual question answering systems are studied \cite{c:513, c:514}, and it is under-studied when it comes to the vision-language joint areas.

\begin{figure*}[htbp]
  \centering
  \includegraphics[width=0.95\linewidth]{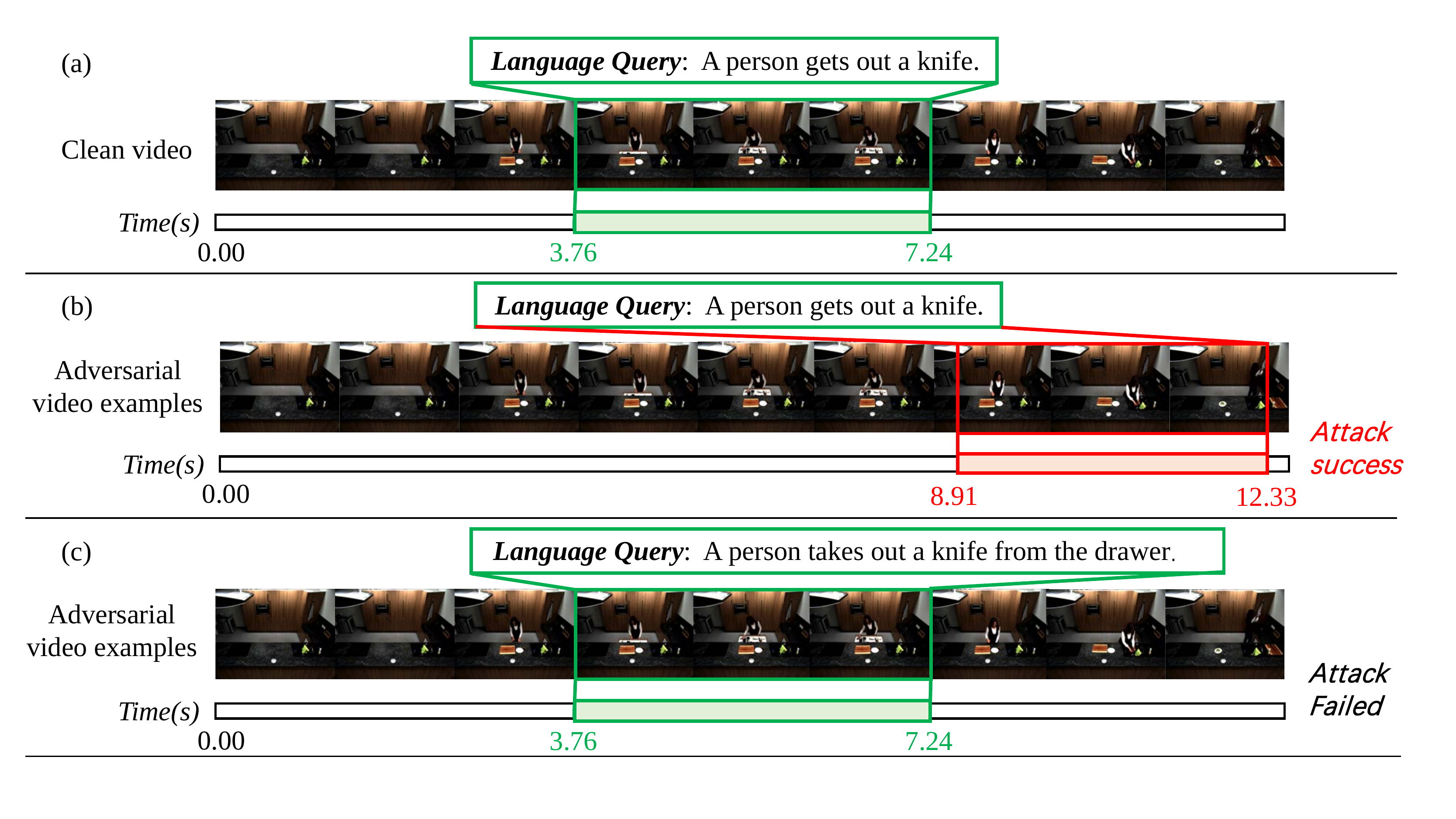}
   \setlength{\abovecaptionskip}{0.cm}
   \vspace{-2em}
   \caption{(a) Given untrimmed video and language query, NLVL model aims at matching the correlated video content with the query and predicting start and end boundary for the video clip. (b) An illustration of adversarial attacks on NLVL models, the adversarial examples will mislead the model to focus on irrelevant segment. (c) Sometimes the adversarial examples generated by the attacker are only effective for one specific query but not for other synonymous sentences.}
   \label{fig:fig1}
\end{figure*}

Still, the robustness of the video-language models like NLVL remains unclear.  Investigating the adversarial vulnerability  of NLVL models encompasses not only the theoretical value of deepening our understanding to this fundamental model characteristic, but also practical significance, especially when considering the widespread applications of NLVL models in security-sensitive scenarios. As a result, it is tempering to ask the following two questions: 
\begin{itemize}[leftmargin=*]
    \item[] \emph{Q1. How do the adversarial attacks on the video and language sides alone affect the overall robustness of the NLVL model?}
    \item[] \emph{Q2. Does there exist even stronger adversarial attacks by explicitly considering the cross-modal nature of the NLVL model?}
\end{itemize}

In this work, we attempt to answer the above questions by making the first effort to comprehensively investigate the adversarial vulnerability of NLVL from both attack and defense perspectives, which covers three facets. In particular, the first two facets target Q1, which serve as the \emph{minor contributions} and stepping stones to the third facet, while the third facet targets Q2 and serves as the \emph{main contribution} of this paper:

\partitle{i) Vulnerability of NLVL to synonymous query sentences in the language side} In practical applications, it is natural to expect the NLVL model to predict the same video clip when queried with various synonymous queries due to varying language habit of different users. Also, synonymous substitutions have been explored in NLP tasks to be devised as an adversarial attack. It serves an alarming sign for NLVL models, which motivates us to evaluate the robustness of NLVL when facing synonymous queries and develop the corresponding synonymous query augmented training strategy as a defense mechanism.

\partitle{ii) Vulnerability of NLVL to the adversarial video perturbation in the vision side} As perhaps the most documented adversarial attack target, image and video tasks-related models have been shown to suffer from adversarial perturbations that are invisible to human eyes but can severely mislead the model prediction. We therefore study the synonymous sentences-oblivion adversarial attack on NLVL, which injects maliciously chosen small perturbations to videos without considering the synonymous query issue in the language side.

\partitle{iii) Vulnerability of NLVL to video-language compound adversarial attack} 
Grounding on the above two facets, we propose a new adversarial attack strategy by considering the cross-modal interactions between the language and video sides, which is called \textbf{S}ynonymous se\textbf{N}tences-awar\textbf{E} \textbf{A}dversarial Attac\textbf{K} on NLVL (\textbf{SNEAK}). As is shown in Figure \ref{fig:fig1}, SNEAK seeks to maximize the adversarial attack capability by choosing the video perturbation that can mislead NLVL over the entire synonymous query sentences set. We also propose a gradient splitting optimization for efficient adversarial perturbation generation, which overcomes the large computational and storage consumption issue. 
In addition, to further enhance the stealthiness of SNEAK, we propose the frame importance-guided pruning mechanism, which suffices to achieve similar attack performance with reduced number of video frames to be perturbed.

\begin{figure}[t]
  \centering
  \includegraphics[width=1.0\linewidth]{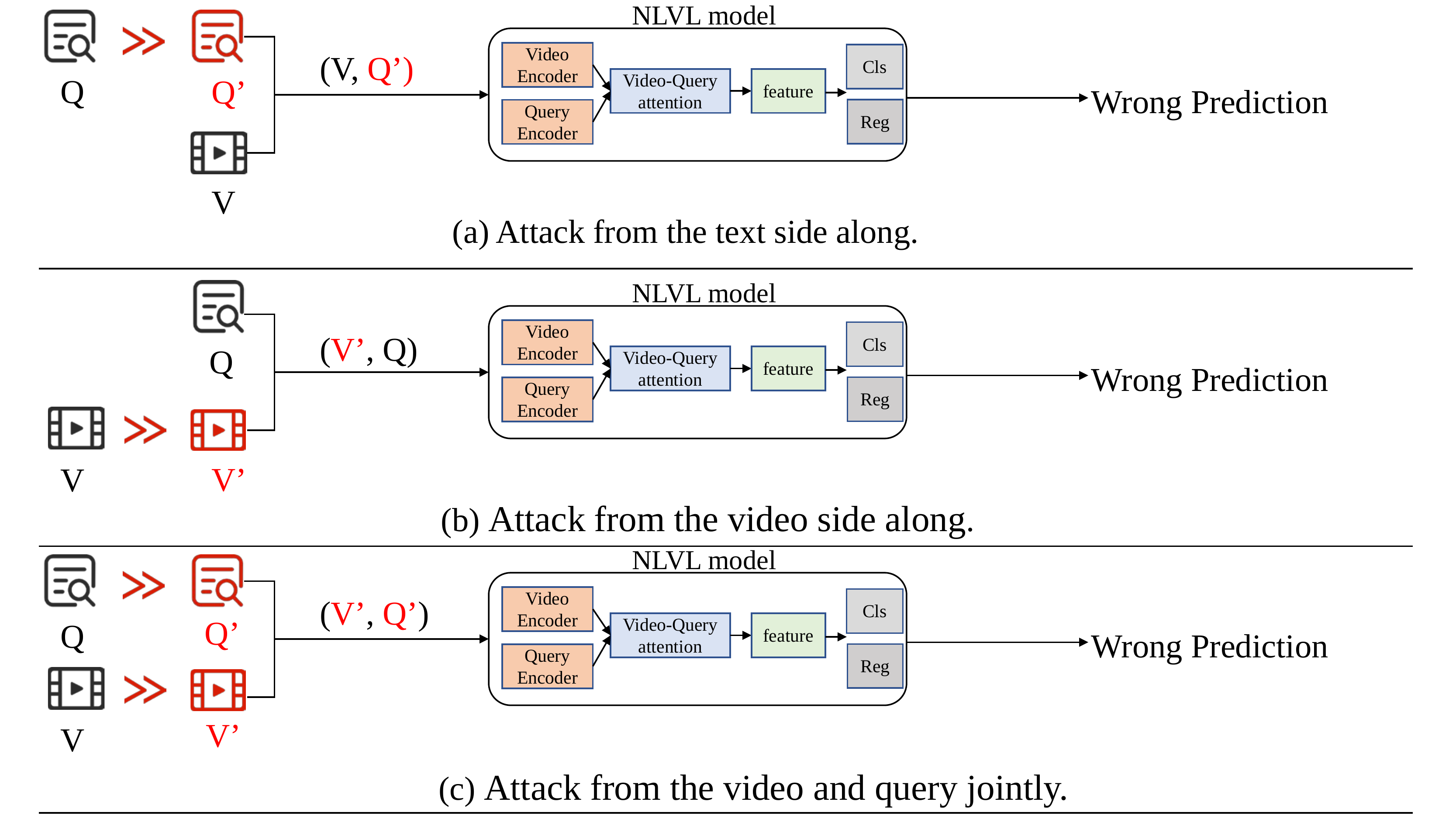}
   \setlength{\abovecaptionskip}{0.cm}
   \vspace{-1em}
   \caption{Three facets of adversarial attacks on NLVL. From top to bottom representing synonymous attack, Synonymous sentences-oblivion attack, and our SNEAK attack, correspondingly.}

   \label{fig:fig2}
\end{figure}

The main contributions can be summarized as follows:
\begin{itemize}
\item To the best of our knowledge, this is the first work to comprehensively investigate the adversarial vulnerability of the NLVL model in the cross-modal video-language area, where we identify three facets of the adversarial vulnerabilities.
\item In order to achieve the cross-modal adversarial attack goal, 
we propose a new synonymous sentences-aware adversarial perturbation attack paradigm, which not only enhances the attack effectiveness on the NLVL model, but also shreds new light on the adversarial attack and defense design on the broader vision-language joint areas. Sneak comes with two new attack algorithm designs for better efficiency and stealthiness.
\item We conduct extensive experiments on the TaCoS and Charades datasets. Experimental results show that our SNEAK attack attains high attack effectiveness. In addition, our SNEAK combined with PSA method maintains high attack performance while enhances stealthiness. The adversarial training-based defense is effective for defending against the new attacks.

\end{itemize}

\section{Related Works}
\label{sec:formatting}

\subsection{Natural Language Video Localization}

NLVL is a typical vision-language joint understanding tasks, which aims to predict the start and end moment boundary that semantically corresponds to a given language query within the untrimmed video, which was firstly introduced in \cite{c:701}. Depending on whether the temporal sliding windows are used \cite{c:519}, existing methods can be grouped into two categories below:

\partitle{Anchor-based NLVL} The anchor-based methods \cite{c:701, c:702, c:703} solve the NLVL by matching the pre-set fixed temporal sliding windows with the natural language query and find the best matching candidate. These early anchor-based studies convert NLVL to a ranking task. Technically, the performance of these methods are largely affected by the number of sliding windows. Since the number of candidate sliding windows is limited, these methods in general cannot guarantee all candidate moments to completely cover the video, which leads to time boundary errors.

\partitle{Anchor-free NLVL} Various anchor-free methods \cite{c:704, c:705, c:706} have been proposed to deal with the above drawback, which directly predict a probability for each frame and choose the frame with the highest probability to be prediction result.Without the need for pre-set limited temporal sliding windows, anchor-free methods build a precise matching mechanism between language query and video. Current work \cite{c:601} regards the NLVL task as a span-based question answering problem by treating the video as a text passage, and and the target temporal locations as the answer span. State-of-the-art work \cite{c:519} utilizes anchor-free methods to generate a group of high-quality candidate video segments with their boundaries, and a classifier is proposed to match the proposals with the sentence by predicting the matching score.

\subsection{Adversarial Attacks and Defenses}

\partitle{Adversarial Attack}
Generating adversarial examples for single modality datasets has been extensively studied recently,  \cite{c:531} propose Fast Gradient Sign Method (FGSM) to generate image adversarial examples utilizing the linear nature of DNN.  Subsequently, a variety of methods such as PGD, C\&W, Deepfool, and SparseFool have been proposed to generate image adversarial examples\cite{c:532}. There are also many studies for adversarial examples against other tasks other than image classification, such as text processing \cite{c:511, DBLP:conf/iclr/ZhaoDS18}, video classification \cite{c:512}, and graph data \cite{c:533}.

Recently, there has been an increasing research interest in generating adversarial examples for multi-modality datasets, such as image question answering systems based on image and text datasets\cite{c:513, c:514, c:534}. However, these studies are all based on images and text modality datasets, while leaving videos and text cross-modality datasets unexplored.

\partitle{Adversarial Defense}
Correspondingly, many defense methods \cite{c:535} have been proposed. Representative approaches include network distillation, adversarial training, adversarial detecting, input reconstruction, network verification and ensemble defenses.

\section{Adversarial Vulnerabilities of NLVL}
\label{sec.adv}
We first formalized the NLVL task, upon which we identify three facets of adversarial vulnerabilities.

\subsection{NLVL Model Formalization}

\begin{definition}
\textbf{(Video-Query Pair)}
Denote the video-query pair as $(\mathcal{V},\Q)$, where $\mathcal{V} \in \mathbb{R}^{T \times C \times H \times W}$ is an untrimmed video with $T$ being the number of frames,
while $C,H,W$ being the corresponding channel, height, and width within each frame; $\Q$ is the text query associating with event in $\mathcal{V}$, which has synonymous sentences set $\mathbb{Q}$. For each video $\mathcal{V}$, an extractor $\E$ is used to extract its visual features $\V=\left\{\mathbf{v}_{i}\right\}_{i=1}^{n}\in\mathbb{R}^{n\times d}$ , where \emph{n} is the number of extracted features and  \emph{$d$} is feature dimension. Similar to visual features, $\Q$ is also mapped into embedding space.
\end{definition}

\begin{definition} \textbf{(Objective formulation of NLVL model)}
    The objective of NLVL model takes the form $\mathcal{L}_{\text span}(J_{\theta}(\V,\Q),(YS,YE))$, where $(\V,\Q))$ is the input video-query pair in the embedding space, $J$ is the NLVL model with trainable model parameters $\theta$, $\mathcal{L}_{\text span}$ is the loss function, $YS$ and  $YE$ are the ground-truth labels of the start and end boundaries, respectively. The loss function takes the form below:
    \begin{equation}
    \begin{small}
\mathcal{L}_{\text {span}}=\frac{1}{2}\left[f_{\mathrm{CE}}\left( PS, YS\right)+f_{\mathrm{CE}}\left(PE, YE\right)\right],
    \end{small}
    \end{equation}
    where $f_{\mathrm{CE}}$ denotes the cross-entropy loss function. 
\end{definition}

\begin{remark}
Video feature extractor is derived from image feature extractor to solve the problem that image feature extractors are only capable to capture spatial information while temporal information is largely or completely neglected. Now the video feature extractors like I3D\cite{DBLP:conf/cvpr/CarreiraZ17},C3D\cite{DBLP:conf/iccv/TranBFTP15}  are widely used as pre-processing step in video related models, for they are able to capture both the spatial pattern and the association between frames and provide abundant temporal message for further process.

\end{remark}

\subsection{Adversarial Vulnerabilities}
\partitle{Adversary assumption}
As the first attempt to investigate the adversary attack on NLVL, we focus on the white box adversarial attack assumption in this paper. In detail, we assume that the adversary knows all the information of the NLVL model details, including the network architecture and model parameters, as well as the synonymous sets of all potential queries, feature extractors for extracting the video and query embeddings. In the model prediction stage, we assume the adversary gets access to the clean video-query pair and the corresponding prediction. 
Equipped with the above information, the adversary is capable to inject imperceptible perturbations to the video frames and replace the query with its synonymous sentences, in the aim to fool the NLVL model to predict incorrect start and end frames that largely deviates from the prediction on the clean input pair.

\subsubsection{Vulnerability of the query side alone}
It is natural for users to feed into the NLSL model with various synonymous query sentences due to varying language habit. Hence, it is of practical importance to evaluate the robustness of NLVL with synonymous query substitutions, especially considering that synonymous substitutions have been devised as an adversarial perturbation to fool NLP models in the pure language tasks. 

For this purpose, we show that some typical NLVL models are not robust to even benign synonymous query substitutions. We utilize WordNet \cite{DBLP:journals/cacm/Miller95} to conduct synonymous substitution, where we find for each query $\Q_i$ a synonymous query set $\mathbb{Q}_i=\{\Q_i,\Q'^{1}_i,\cdots,\Q'^{5}_i\}$, i.e., containing the original query sentence and 5 different synonymous substituted queries. Each of the substituted query has at least 2 words swapped.

Our experiment results in Section \ref{sec.experiment} (e.g., the first row of Table \ref{Table bound} reveal the vulnerability of NLVL subjected to even benign synonymous query substitution.

\subsubsection{Vulnerability of the video side alone}
It is well-recognized that imperceptible perturbations to image and video can fool DNN models. We evaluate such adversarial perturbations by the following attack, which does not consider the interplay with the language query side.
\begin{definition}\label{def.obli}
\textbf{(Synonymous Sentences-Oblivion Adversarial Attack on NLVL)}
Denote the loss of NLVL by $\mathcal{L}_{\tt span}$, the targeted NLVL model by $J_{\theta}$. For an input video-query pair by $(\V,\Q)$ with ground-truth predict $(YS,YE)$, the synonymous sentences-oblivion adversarial attack on NLVL perturbs the video $\V$ to $\V_{obl}$ by $\delta _{obl}$ designated by 
\begin{gather}
    \bm{\delta}_{obl} = \argmax _{\bm{\delta}} \mathcal{L}_{\tt span}(J(\V+\bm{\delta},\Q),(YS,YE)),\\
    s.t.,~~\|\bm{\delta}\|_2 \leq B.
\end{gather}
\end{definition}
The above $\bm{\delta}_{obl}$ can be obtained by projected gradient descent (PGD) method following the adversarial attack literature. Our experiment results in Section \ref{sec.experiment} confirms that NLVL is also vulnerable to such video adversarial perturbation without considering synonymous sentences.

\subsubsection{Cross-modal vulnerability of video and query sides jointly}
The cross-modal nature of NLVL intrigues us to ask whether an even stronger adversarial attack exists by explicitly taking consideration of both the video and query sides. To answer this question, we take inspiration from the above two vulnerabilities to come up with a new cross-modal adversarial attack on NLVL by proposing a new attack paradigm as presented in the next section.

\section{Proposed Adversarial Attack on NLVL}
In this section, we first propose our new \textbf{SNEAK}: \textbf{S}ynonymous se\textbf{N}tences-awar\textbf{E} \textbf{A}dversarial Attac\textbf{K} on NLVL, which captures the cross-modal video-query interaction in adversarial perturbation design, Figure \ref{fig:fig2} reveals the difference of three types of attacks. Then, we produce a further frame gradient importance-guided pruning strategy to reduce the number of frames to be perturbed in a victim video in aim to further enhance the stealthy of the adversarial attack. 

\subsection{SNEAK Attack Formulation}
\label{subsec.sneak}
\begin{definition} \label{def.aware}
\textbf{(SNEAK: Synonymous Sentences-Aware Adversarial Attack on NLVL))}
 Denote the loss of NLVL by $\mathcal{L}_{\tt span}$, the targeted NLVL model by $J$. For an input video-query pair by $(\V,\Q)$ with ground-truth predict $(YS,YE)$, denote the synonymous query sentences set of $\Q$ by $\mathbb{Q}$. The synonymous sentences-aware adversarial attack on NLVL adversarial perturbs the video $\V$ to $\V_{awr}$ by $\delta _{awr}$ is designated: 
\begin{gather}
    \bm{\delta}_{awr}^{best} = \argmax _{\bm{\delta}} \min_{\Q'\in\mathbb{Q}}\mathcal{L}_{\tt span}(J(\V+\bm{\delta},\Q'),(YS,YE)),\label{eq.min}\\ 
    s.t., \|\bm{\delta}\|_2 \leq B.
\end{gather}
\end{definition}
\begin{remark}
Compared with the previous synonymous sentences-oblivion adversarial attack in Definition \ref{def.obli} that obtains the video perturbation without considering the synonymous queries, Sneak in Definition \ref{def.aware} seeks to find the video perturbation that deteriorates the prediction for all synonymous queries. In particular, eq.(\ref{eq.min}) achieves this goal by formulating the objective function of the video perturbation $\delta_{awr}^{min}$ as a max-min optimization problem, which can be interpreted as to find $\delta_{awr}^{min}$ that maximizes the NLVL model loss even on the most well-predicted synonymous query.
\end{remark}
\partitle{Key algorithm step for obtaining $\bm{\delta}_{awr}^{min}$} Sneak can be solved via an iterative algorithm based on the projected gradient ascent algorithm, which has the key per-iteration step as depicted below
\begin{equation*}
\begin{small}
    \bm{\delta}[t+1] = 
    \mathcal{P}(\bm{\delta}[t] + \lambda[t]\nabla_{\bm{\delta}} \mathcal{L}_{\tt span}(J(\V+\bm{\delta},\Q^{\dagger}[t]),(YS,YE))),
\end{small}
\end{equation*}
where $\lambda [t]$ is the learning rate, $\nabla_{\bm{\delta}}\mathcal{L}_{\tt span}(J(\cdot))$ is the partial gradient taken with respect to $\bm{\delta}$, $\Q^{\dagger}[t])$ is the synonymous sentence with the smallest loss at iteration $t$ that corresponds to the most-well predicted synonymous query at the current $\bm{\delta}[t]$. In particular, $\mathcal{P}(\x)$ projects vector $\x$ onto the $B$-$\ell _2$ norm bounded ball by $\mathcal{P}(\x) = \frac{\x}{\|\x\|_2}\cdot\min\{B, \|\x\|_2\}$. 

In the following, we also investigate two variants of Sneak by relaxing eq.(\ref{eq.min}) alternatively as below:

\partitle{``Average'' variant} The ``average'' variant of adversarial video perturbation $\bm{\delta}_{awr}^{avg}$ is obtained by relaxing the NLVL loss from on the most well-predicted query sentence to on the average synonymous sentences: 
\begin{equation*}
\begin{small}
    \bm{\delta}_{awr}^{avg} = \argmax _{\bm{\delta}} \frac{1}{|\mathbb{Q}|}\sum _{\Q' \in\mathbb{Q}}\mathcal{L}_{\tt span}(J(\V+\bm{\delta},\Q'),(YS,YE)).
\end{small}
\end{equation*}
Perturbation $\bm{\delta}_{awr}^{avg}$ has the following per-iteration step at iteration $t$: 
\begin{equation*}
    \begin{split}
     &\bm{\delta}[t+1] = \\
     &\mathcal{P}(\bm{\delta}[t] + \lambda[t]\frac{1}{|\mathbb{Q}|}\sum_{\Q'\in\mathbb{Q}}\nabla_{\bm{\delta}} \mathcal{L}_{\tt span}(J(\V+\bm{\delta},\Q',(YS,YE))),
    \end{split}
\end{equation*}
where $\Q'$ can also be a randomly sampled subset of synonymous sentences rather than the entire $\mathbb{Q}$, which motivates the following variant.

\partitle{``Random'' variant:} The ``random'' variant of adversarial video perturbation  $\bm{\delta}_{awr}^{rand}$ is obtained by iterating the following step:
\begin{equation*}
\begin{small}
     \bm{\delta}[t+1] = 
     \mathcal{P}(\bm{\delta}[t] + \lambda[t]\nabla_{\bm{\delta}} \mathcal{L}_{\tt span}(J(\V+\bm{\delta},\Q[t]),(YS,YE))),
\end{small}
\end{equation*}
where $\Q[t]$ is a randomly sampled synonymous query. 
\begin{remark}
The above three variants manifest different levels of trade-off between computational cost and attack strength. The ``best'' and the ``average'' variants respectively requires extra feed-forward and back-propagation evaluations over all synonymous sentences, while brings more computation during adversarial perturbation generation, while the ``random'' variant tends to sacrifice certain attack strength in exchange for less computation. Our empirical results in Section \ref{sec.experiment} will illustrate the trade-off.
\end{remark}

\subsection{Gradient Splitting Optimization}
Due to the large-scale video and text datasets, feature extractors and cross-modal interaction networks are usually trained separately in current NLVL models. If one stacks networks up and trains the stacked networks end-to-end, the model parameters will be incredibly large, which requires massive computational overhead and GPU memory for generating adversarial perturbations when obtaining the gradient from the video input. For this reason, we propose the gradient splitting optimization method which adds the $\ell_2$-norm constrained perturbations to the video features instead of the original video pixels. Afterwards, we use video feature extractor network to generate the video adversarial examples through gradient descent-based fitting method. At first glance, such splitted optimization tends to generate suboptimal video perturbations, which can lead to less effective adversarial attack. However, our empirical results show that it is capable to lead to sufficient NLVL performance drop and hard to detect pixel perturbations. As a result, our gradient splitting optimization mitigates the huge computational overheads, while generating powerful enough adversarial perturbations.

Having obtained $\bm{\delta}_{awr}$ from Section \ref{subsec.sneak} and knowing the feature extractor $\E$ used by target model, we reverse the step of extracting video features: adding pixel space perturbations $\bm{\delta}_{ps}$ on clean video $\mathcal{V}$ as extractor input $(\mathcal{V}+\bm{\delta}_{ps})$ and obtain video feature:
\begin{equation}
\begin{small}
\V_{adv}=\E((\mathcal{V} + \bm{\delta}_{ps})),
\end{small}
\end{equation}
we choose the optimization target to be Mean Square Error ($MSE$) loss $\mathcal{L}_{\tt vp}$ between $\V_{adv}$ and target output $\bm{\delta}_{awr}$, and therefore through iterations gradually approaches:
\begin{equation}
\begin{small}
    \min _{\bm{\delta}_{ps}}\mathcal{L}_{\tt vp}=MSE(\V_{adv},\emph{\V}+\bm{\delta}_{awr}).
\end{small}
\end{equation}
Finally, we get the perturbations $\bm{\delta}_{ps}$ on pixel space of the video $\mathcal{V}$.

\subsection{Frame Gradient Importance-guided Perturbation Pruning}
In order to reduce the number of frames that need to be disturbed while still ensuring the effect of the attack, we use a pruning-based method to filter out part of $\bm{\delta}$ that has little influence on fooling the NLVL model. This method is inspired by the model compression algorithm based on pruning, the difference is that the existing pruning method is to compress the model, while we are pruning the noise with less effect on the prediction result in order to reduce the number of frames to be perturbed. This method not only reduces the time for training adversarial samples, meanwhile it makes our attack more stealthy. We then formulate the indices to be perturbed as follows:
\begin{equation}
\begin{small}
    {\mathrm{T}=\mathrm{TopK}\left(\operatorname{argsort}\left(\| \bm{\delta}_{awr}^{(\cdot)}\|_{2} )\right)\right)},
\end{small}
\end{equation}
where $\mathbf{T} \in\{\mathbf{0}, \mathbf{1}\}^{n\times d }$ is the temporal mask of feature. We let $\Theta=\{1,2, \ldots, T\}$  be the set of feature indices, $\Phi$ be a subset within $\Theta$, and $\Psi=\Theta-\Phi$. If $t \in \Phi$, $T_{i}=0$, and if $t \in \Psi, T_{i}=1$, where $T_{i} \in\{\mathbf{0}, \mathbf{1}\}^{d}$ is the $i$-th feature in $V$. 

Since most of the video feature extractors, they can only map several clips into single feature, but still lacking the ability to reflect the relationship of one frame and another remote one. Therefore each of the extracted feature is independent to one another. Thus with this prerequisite, we are able to select clips of frames and manipulate them, meanwhile still be confident not bringing extra influence to neighbouring frames, which is designed to stay unpolluted. In this way, we enforce the computed perturbations to be added only on the selected video features. For example, in order to obtain the pruning-based ``best'' adversarial perturbation, the objective function is modified as follows:

\begin{gather}
    \bm{\delta}_{awr}^{best} = \argmax _{\bm{\delta}} \min_{\Q'\in\mathbb{Q}}\mathcal{L}_{\tt span}(J(\V+T\cdot\bm{\delta},\Q'),(YS,YE)),\label{eq.min}\\ 
    s.t., \|T\cdot \bm{\delta}\|_2 \leq B.
\end{gather}

Again, the generation of adversarial example includes three steps. The first step is to add adversarial perturbations on the video feature; the second step is to prune the added perturbation, and the final step is to use the video feature extractor $\E$ of the NLVL model to fit the pruned adversarial features based on optimized methods to get the pixel adversarial example.

\subsection{Defense with Adversarial Training}
As a complimentary aspect, we study the defence against the SNEAK attack with adversarial training, which has the following formulation:
\begin{equation}
\begin{small}
    \min _{\theta} \expect \Big{[} \max_{\bm{\delta}_{awr}^{(\cdot)}} \mathcal{L}_{span}(J_{\theta}(\V+\bm{\delta}),\Q'),(YS,YE) \Big{]},
\end{small}
\end{equation}
where the expectation is taken with respect to the data distribution of video-query pairs, $\bm{\delta}_{awr}^{(\cdot)}$ is the synonymous sentences-aware video perturbation that can be generated with one of the variants in Section \ref{subsec.sneak}, and $\Q'$ is a synonymous sentence in $\mathbb{Q}$. 

During adversarial training, we alternatively select the videos from clean video $\V$ and perturbed video $(\V+\bm{\delta})$ between training epochs, in order to encourage the model to learn from both clean and adversarial video samples. As for the query sentence selection, we follow the ``random'' variant to random sample $\Q' \in \mathbb{Q}$. In particular, the per-sample gradient takes the form below:
\begin{equation}
    \left\{
            \begin{array}{lr}
                 \nabla _{\theta}\mathcal{L}_{\tt span}(J_{\theta}(\V,\Q'),(YS,YE)), &\text{if epoch is odd}\\
                 \nabla _{\theta}\mathcal{L}_{\tt span}(J_{\theta}(\V+\bm{\delta}_{awr}^{rand},\Q'),(YS,YE)), &\text{if epoch is even}
            \end{array}
            \right\}
\end{equation}
where $\nabla _{\theta}\mathcal{L}_{\tt span}(J_{\theta}(\cdot))$ takes gradient with respect to the NLVL model parameters $\theta$.

\section{Experiments}
\label{sec.experiment}
We evaluate the effectiveness of the proposed adversarial attacks and defense on one of the state-of-the-art NLVL model with two real-world datasets commonly utilized in NLVL literature. Due to space limit, we relegate the detailed experiment settings (e.g., hyperparameter selections, data pre-processing) and more experiment results (e.g., results on different NLVL models, on additional datasets, more visualization) to the supplementary material.

\begin{figure}[htbp]
  \centering
  \includegraphics[width=1.0\linewidth]{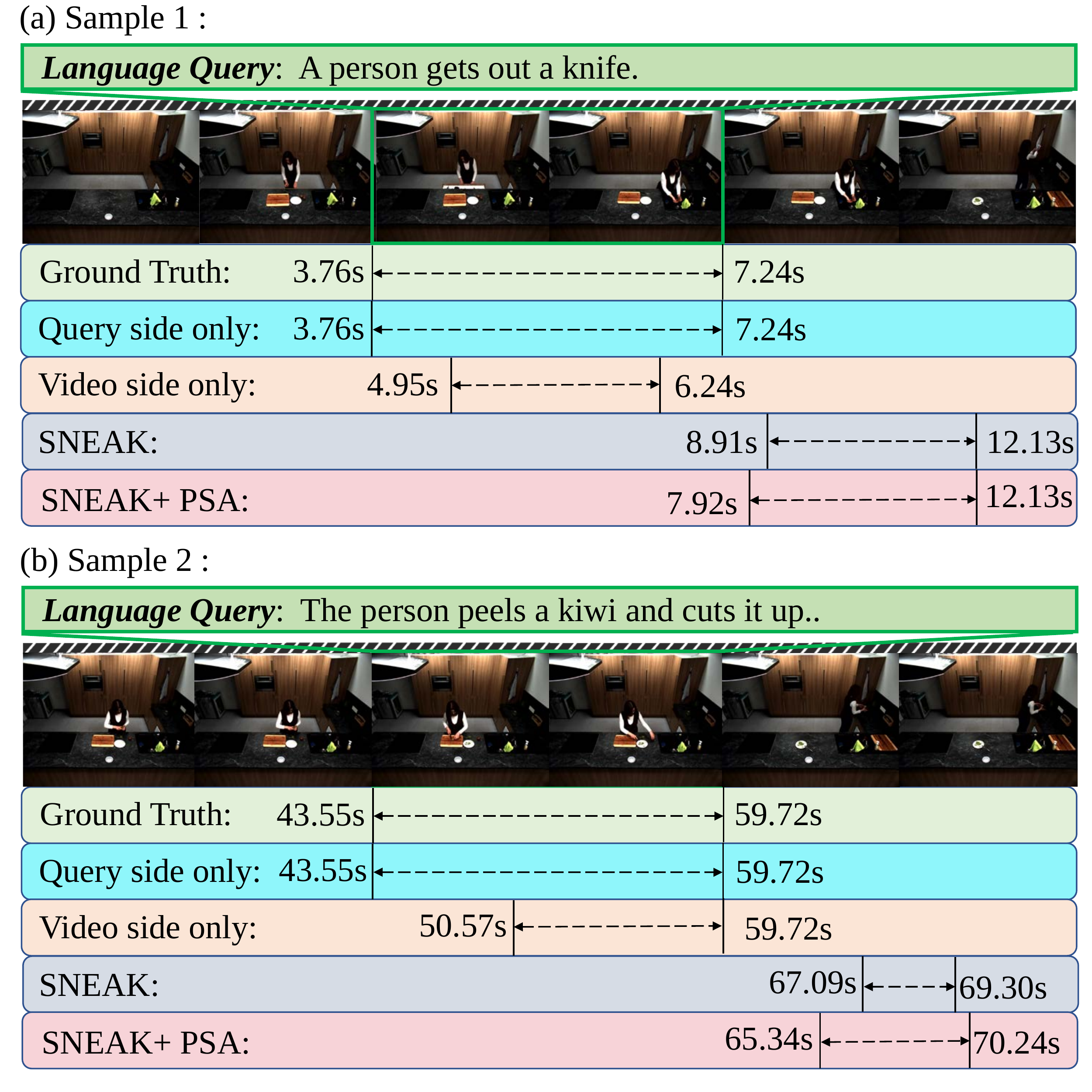}
  
   \setlength{\abovecaptionskip}{0.cm}
   \caption{Qualitative attacking results sampled from TACoS Datasets. When tested with synonymous query, both query side alone and video side alone methods fail to attack. In addition, SNEAK has better attack performance than SNEAK+PSA, while SNEAK+PSA method is more stealthy due to less perturbed frames.}
   \label{fig:Qualitative}
      \vspace{-1em}
\end{figure}

\subsection{Experiment Setup}
\partitle{Baseline models}
We utilize VSLNet \cite{c:601} as an exemplary NLVL model for evaluation, which demonstrates state-of-the-art performance. 
VSLNet is  a typical anchor-free NLVL model, i.e., it does not rely on candidate video segments and capable to perform the prediction directly on the entire queried video. The main network structure of VSLNet consists of two shared feature encoders, an context-query attention module, a query-guided highlighting module, and a conditioned span predictor.

\partitle{Datasets}
Two mainstream datasets are considered, for which we follow the same setting with the original VSLNet \cite{c:601} for best model performance without attack.

\partitle{TACoS dataset} It is based on MPII Cooking Composite Activities dataset \cite{c:602}, where 10,146 samples as are used for training and 4,083 samples are used for testing.

\partitle{Charades-STA dataset} It contains videos of daily indoor activities based on Charades dataset (Sigurdsson et al. 2016), where 12,408 samples are used for training  and 3720 are used for testing. 

\partitle{Evaluation metrics}
Following NLVL literature, Intersection over Union (IoU) is adopted as the evaluation metric, we use $mIoU$ as our evaluation metric, which is the average IoU over all testing samples given by
\begin{equation}
\begin{small}
 mIoU=\frac{1}{K}\sum_{k=1}^{K} IoU_k,
\end{small}
\end{equation}
where $K$ is the total number of the testing set.

\subsection{SNEAK Adversarial Attack Results}\label{vid attack}

\begin{figure}[t]
  \centering
  \includegraphics[width=1.0\linewidth]{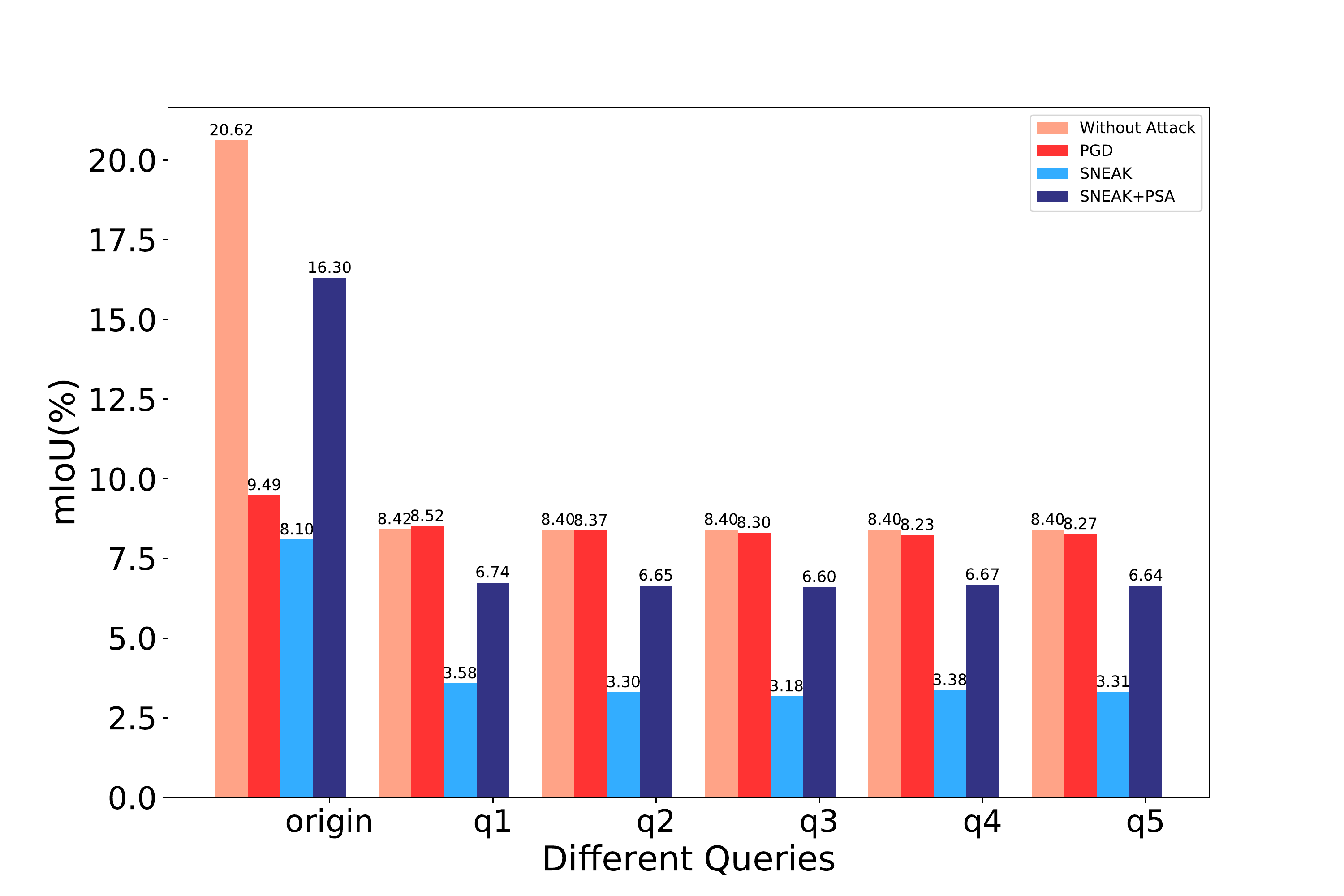}
   \setlength{\abovecaptionskip}{0.cm}
      \vspace{-1.5em}
   \caption{Attacking results of different methods.}
   \label{fig:onecol}
   \vspace{-1.5em}
\end{figure}

In this part, we evaluate the attack effectiveness of SNEAK, where the results of the clean input and the video perturbed by the synonymous sentences-oblivion adversarial attack obtained by PGD in Section 3.2.2 are compared. As shown in Figure \ref{fig:onecol}, PGD manages to deceive the model, which achieves sufficient prediction performance dropping on the original query sentence. However,  the attacking only induces a slight decrease on synonym queries, for there might be a mismatch between PGD generated perturbation, video and synonym queries, causing an unsuccessful attack. 

Meanwhile the top row of Table \ref{Table bound} indicates that only applying synonymous substitution achieves great attack performance, similar to only applying the PGD noise. However, $\delta$ generated by SNEAK displays its significant high capacity on attacking target NLVL models, the attack out performed PGD method on original query, while noticeably more effective on attacking video with synonym substitution queries than applying previous attacking methods alone, for SNEAK takes a greater query set into consideration when obtaining noise.  The overall mIoU of predictions drop from above 8\% to around 3\%, which is only 1/8 of original model performance. We then propose that, for multi-modal or cross modal attacks, only perturbing single modal input will gain success, but manipulating all modal inputs simultaneously will lead to more promising attacking result.\color{black}

\subsection{SNEAK with PSA Enhancement Results}

\begin{figure*}[htbp]
  \label{fig:PTBloss}
    \begin{minipage}[t]{0.5\columnwidth}
      \flushleft
        \includegraphics[width=1.125\columnwidth]{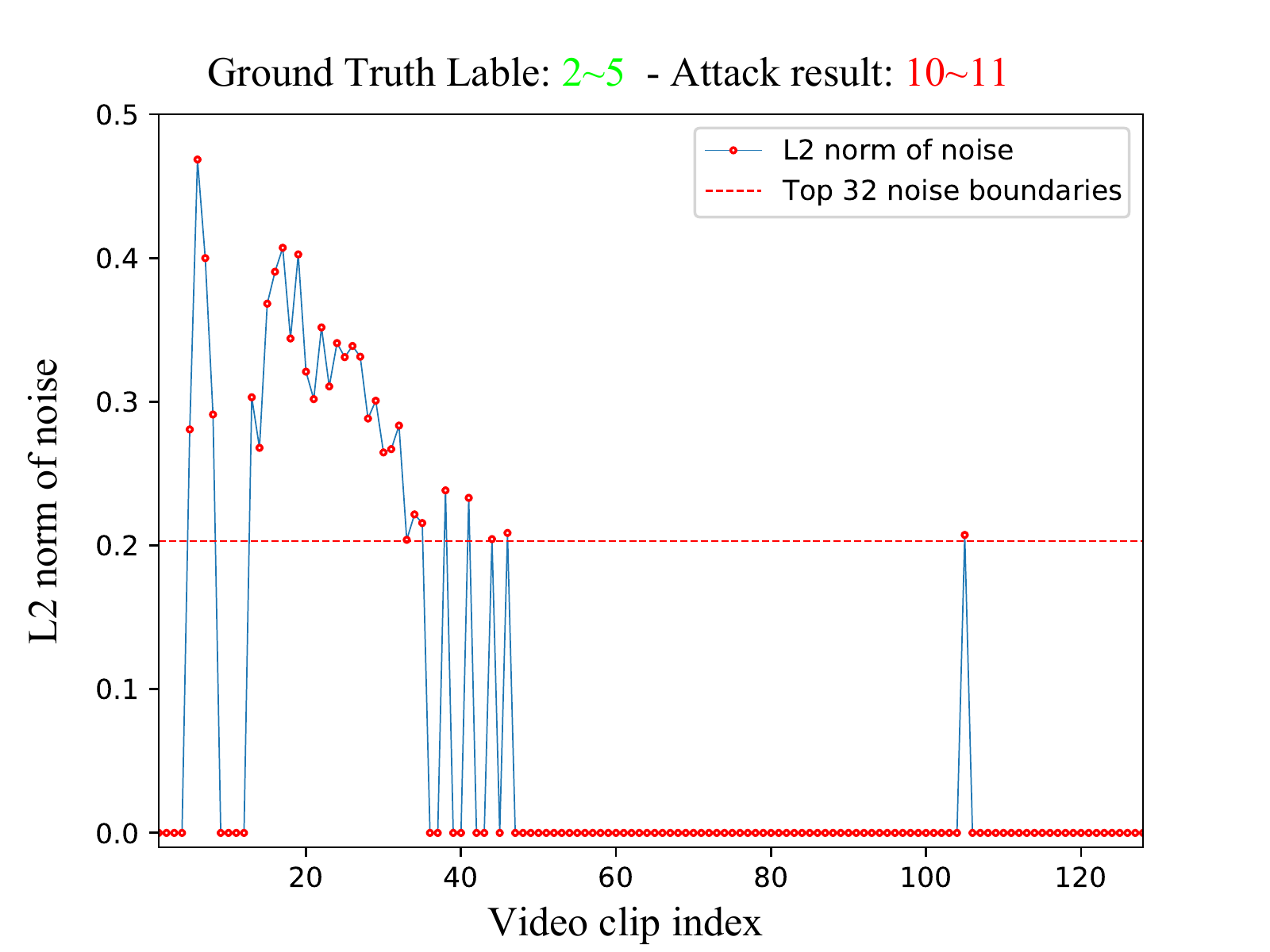}
    \end{minipage}
    \begin{minipage}[t]{0.5\columnwidth}
    \flushright 
        \includegraphics[width=1.125\columnwidth]{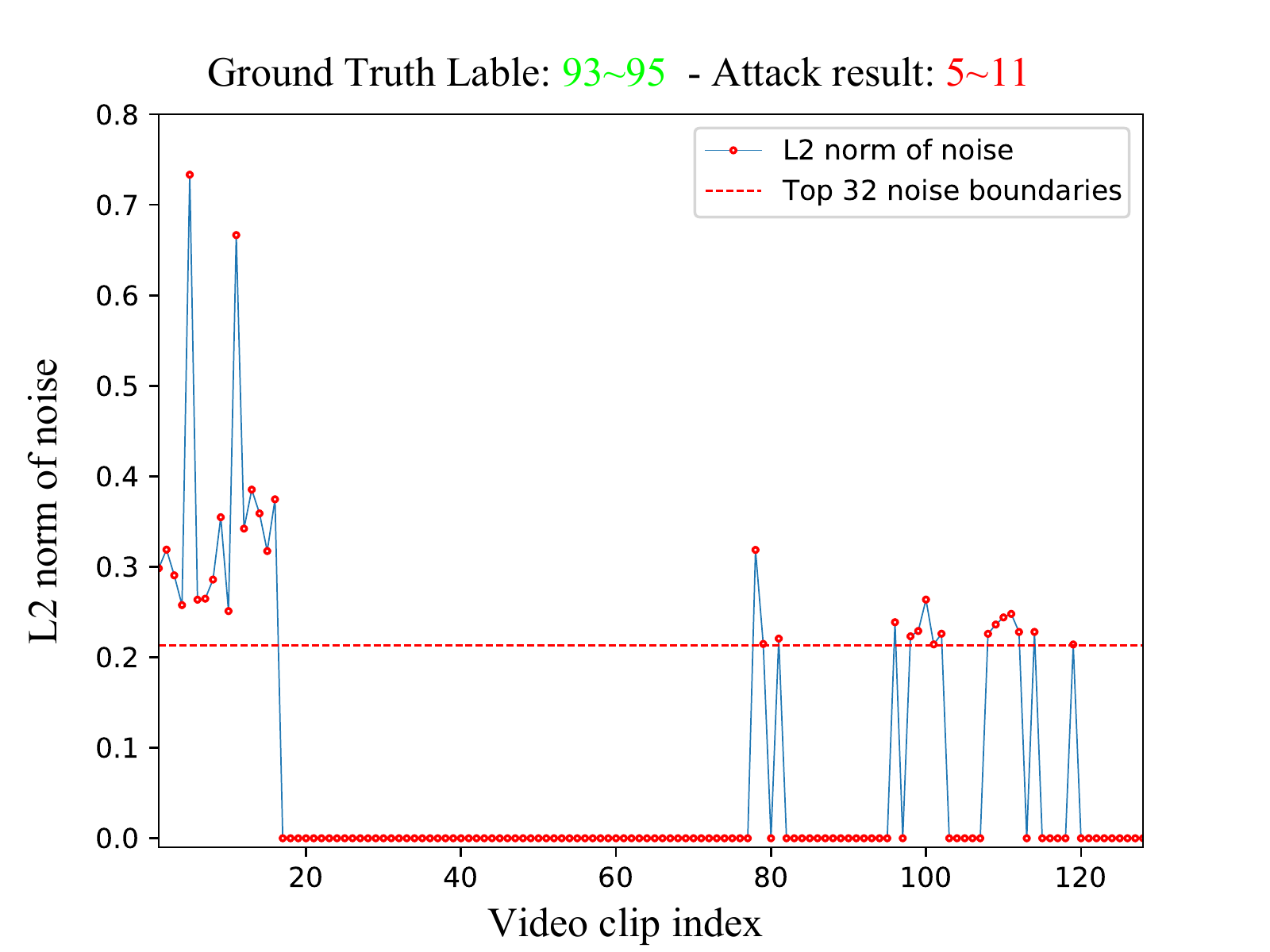}
    \end{minipage}
    \begin{minipage}[t]{0.5\columnwidth}
    \flushright 
        \includegraphics[width=1.125\columnwidth]{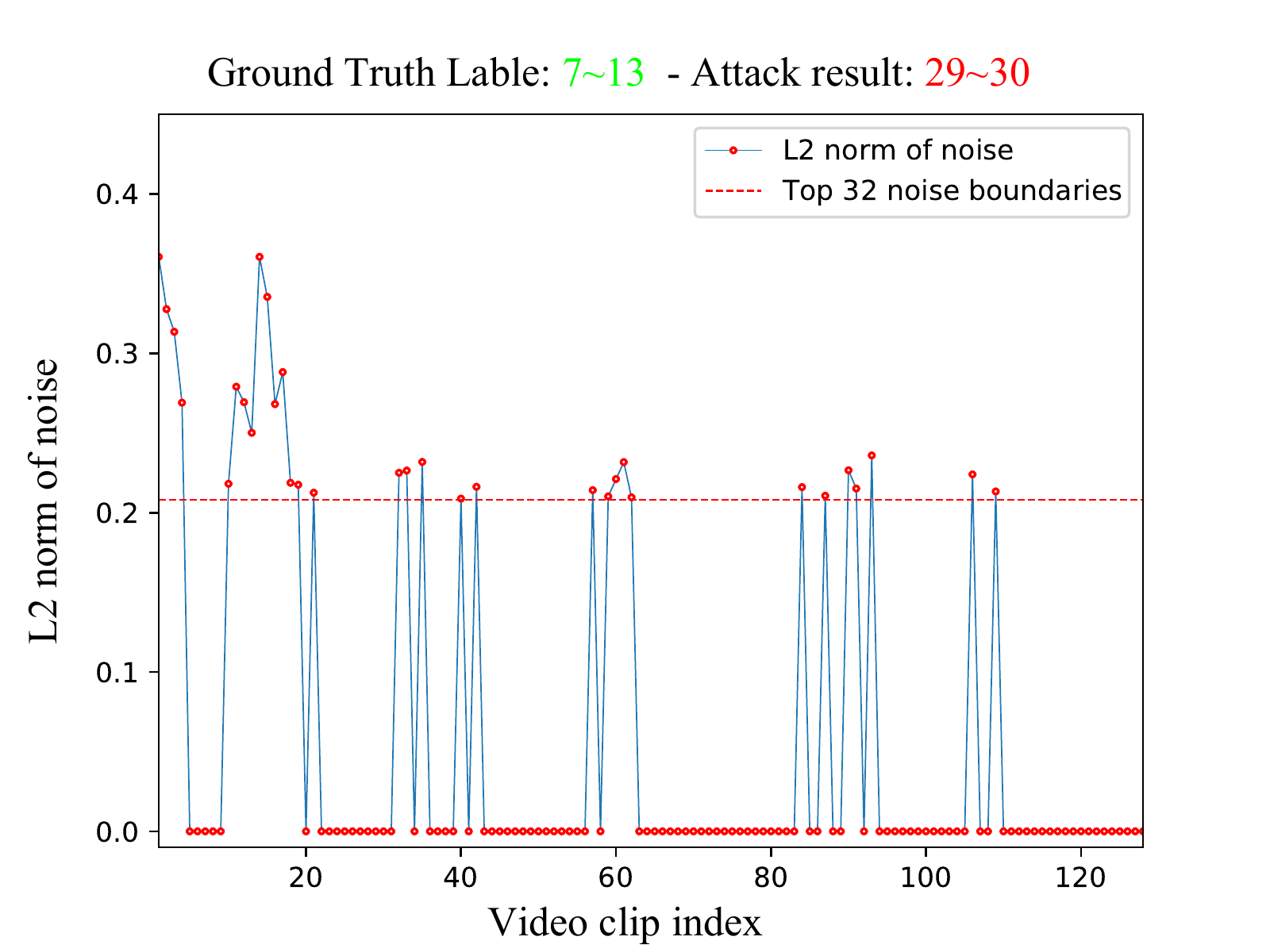}
    \end{minipage}
    \begin{minipage}[t]{0.5\columnwidth}
    \flushright 
        \includegraphics[width=1.125\columnwidth]{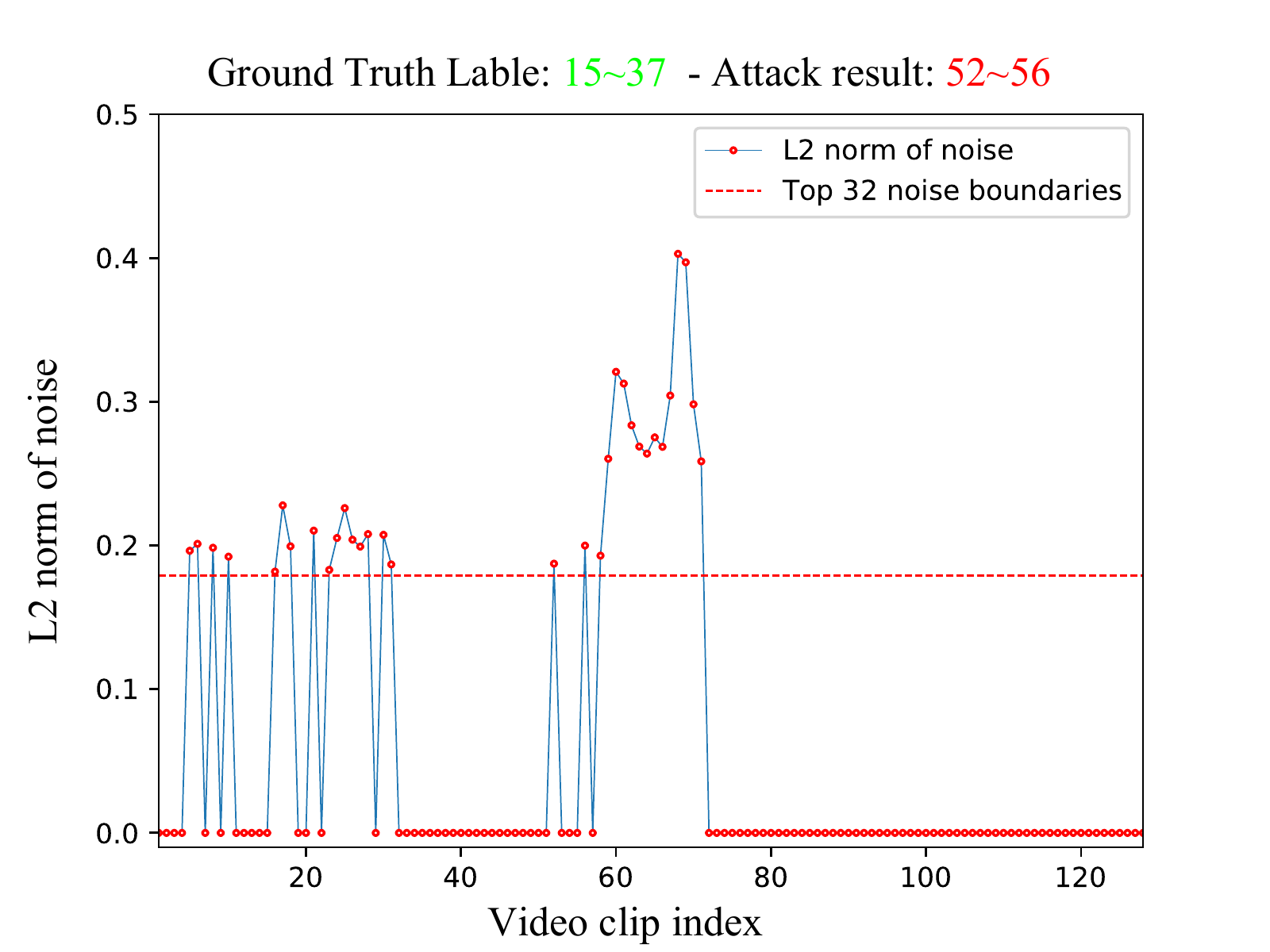}
    \end{minipage}
  \label{fig:ComarisonPTB}
  \setlength{\abovecaptionskip}{0.cm}
  \caption{\ Examples showing the scale of SNEAK+PSA generated noise on each clip of the video feature $\V$, as well as ground truth label in green text, prediction result after attack in red text. Most of the noise injected roughly cover both ground truth label and predict result after attack, indicating perturbation around these clips has the most significant impact to mislead the NLVL model.}
  \vspace{-1.5em}
\end{figure*}

\begin{table}[htbp]
    \setlength{\abovecaptionskip}{0.cm}
    \caption{Comparison of mean IoU ($\%$) for SNEAK+PSA methods with different bound, pruning size=25\%}
  \label{tab:my_label}
  \centering
  \setlength{\tabcolsep}{1mm}{
  \begin{tabular}{c|c|c|c|c|c|c}
        \hline
        \multirow{2}*{bound} & \multicolumn{6}{c}{Query} \\
        \cline{2-7} 
                  & origin                & $q_1$        & $q_2$          & $q_3$             &$q_4$       & $q_5$ \\ 
        \hline
        \hline
        without attack    & 24.110             & 8.419  & 8.398        & 8.398   & 8.404           &8.405     \\
        5  &  16.297            & 6.738           & 6.651          & 6.604            &6.672                 & 6.640 \\ 
        7.5 &  10.380             & 4.430          & 4.100           & 3.990           &3.930                  & 4.100 \\
        10  & 10.330     &4.200           & 4.020  & 4.030  &4.130         & 4.110 \\ 
        \hline 
  \end{tabular}
  }\label{Table bound}
     \vspace{-1.5em}
\end{table}
\begin{table}[!h]
    \setlength{\abovecaptionskip}{0.cm}
    \caption{Comparison of mean IoU ($\%$) for SNEAK+PSA methods with different pruning size, b=7.5}
  \label{tab:my_label}
  \centering
  \setlength{\tabcolsep}{1mm}{
  \begin{tabular}{c|c|c|c|c|c|c}
        \hline
        \multirow{2}*{noise size(\%)} & \multicolumn{6}{c}{Query} \\
        \cline{2-7} 
                  & origin                & $q_1$        & $q_2$          & $q_3$             &$q_4$       & $q_5$ \\ 
        \hline
        \hline
        without attack    & 24.110             & 8.419  & 8.398        & 8.398   & 8.404           &8.405     \\
        12.5  &    12.390            & 5.590           & 5.360          & 5.630          &5.300                & 5.340 \\ 
        25 &  10.380             & 4.430         & 4.100          & 3.930           &3.930                  & 4.100 \\
        50  & 7.340     &2.770          & 2.650  & 2.340  &2.480         & 2.640 \\ 
        \hline 
  \end{tabular}
  }\label{Table size}
\end{table}

\begin{figure}[htbp]
  \label{fig:PTBloss}
    \begin{minipage}[t]{0.47\columnwidth}
      \flushleft
        \includegraphics[width=1.15\columnwidth]{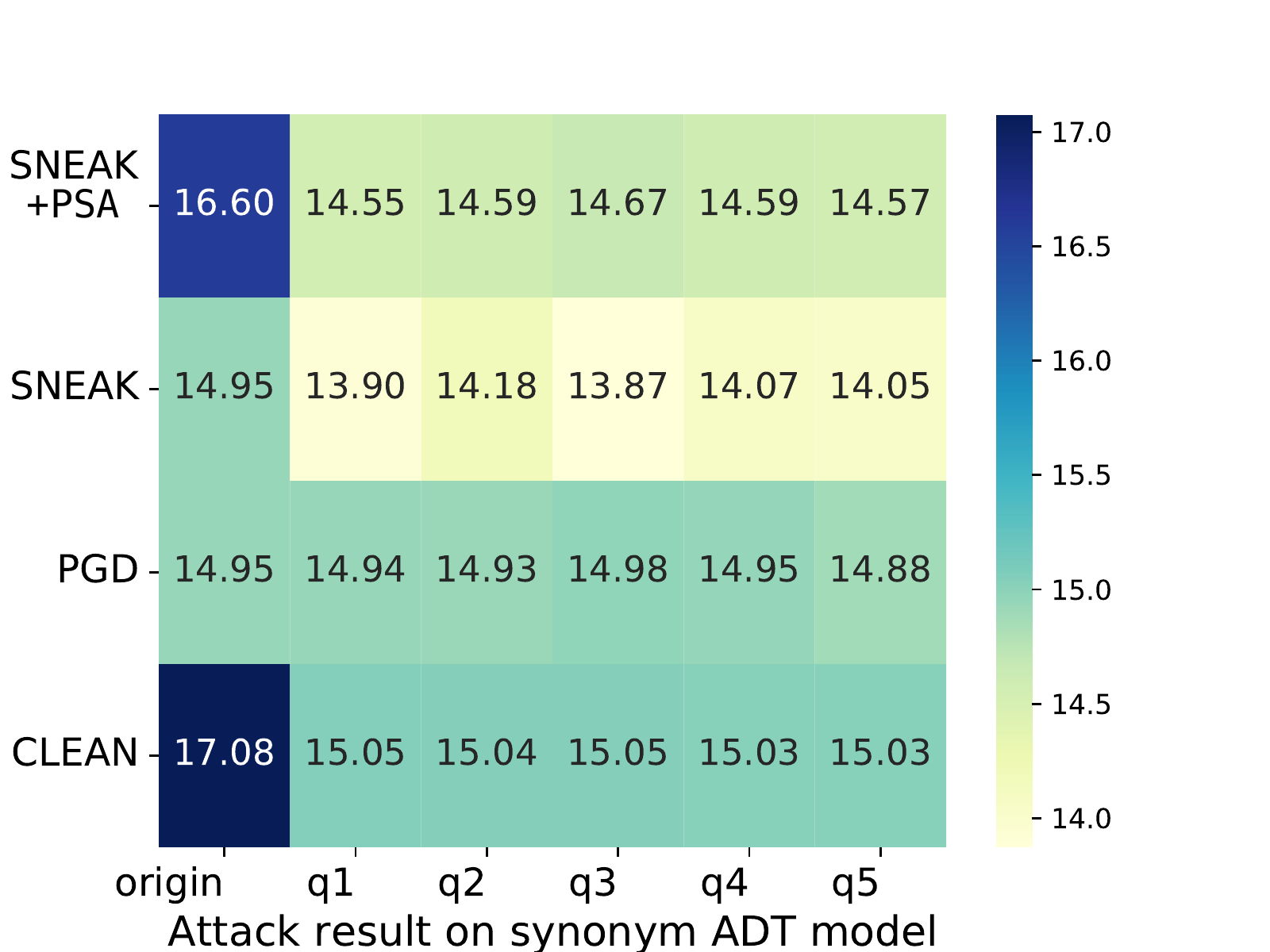}
    \end{minipage}
    \begin{minipage}[t]{0.47\columnwidth}
    \flushright 
        \includegraphics[width=1.15\columnwidth]{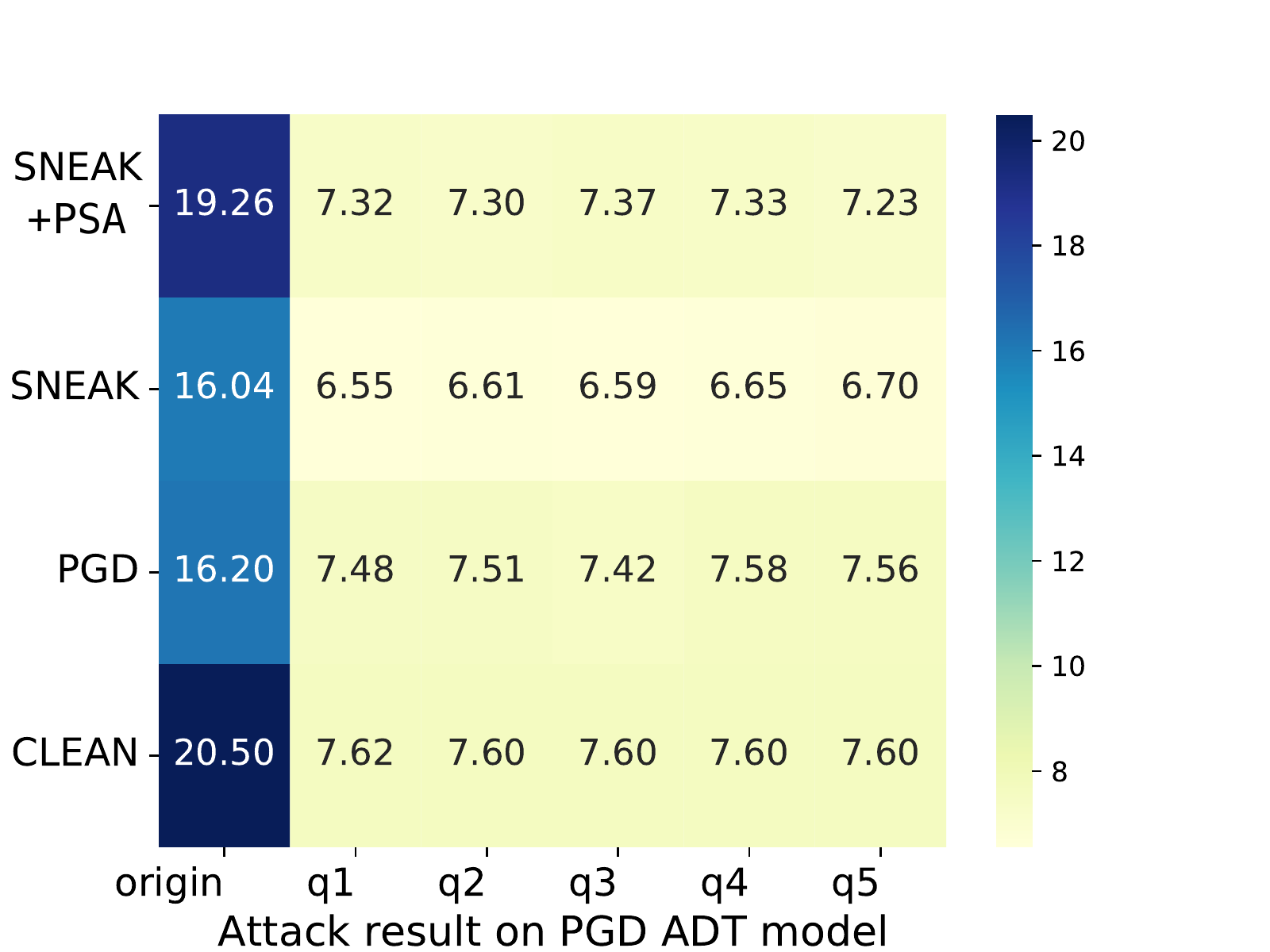}
    \end{minipage}
       \vspace{-1em}
\end{figure}
\begin{figure}[htbp]
  \label{fig:PTBloss}
    \begin{minipage}[t]{0.47\columnwidth}
      \flushleft
        \includegraphics[width=1.15\columnwidth]{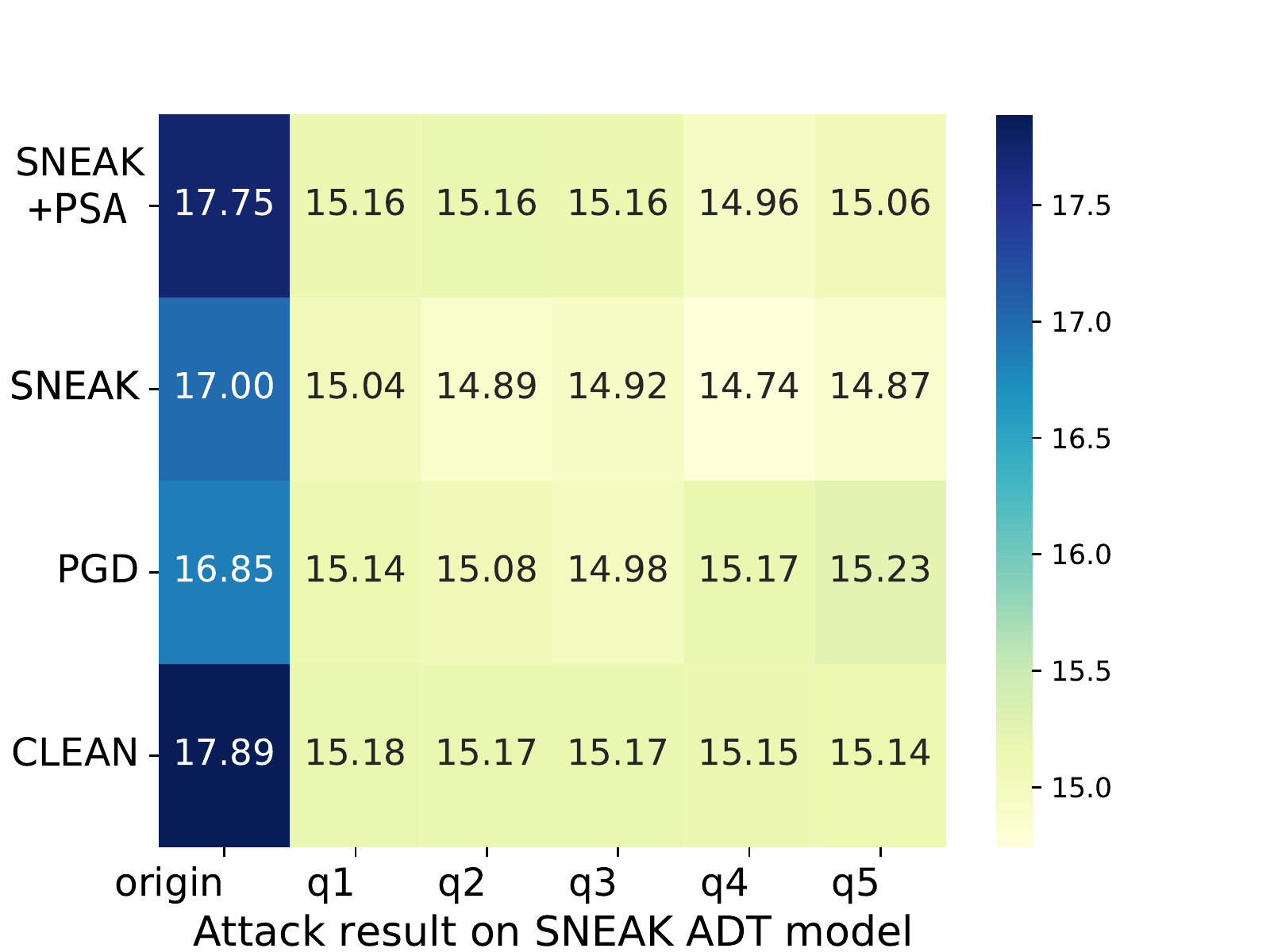}
    \end{minipage}
    \begin{minipage}[t]{0.47\columnwidth}
    \flushright 
        \includegraphics[width=1.15\columnwidth]{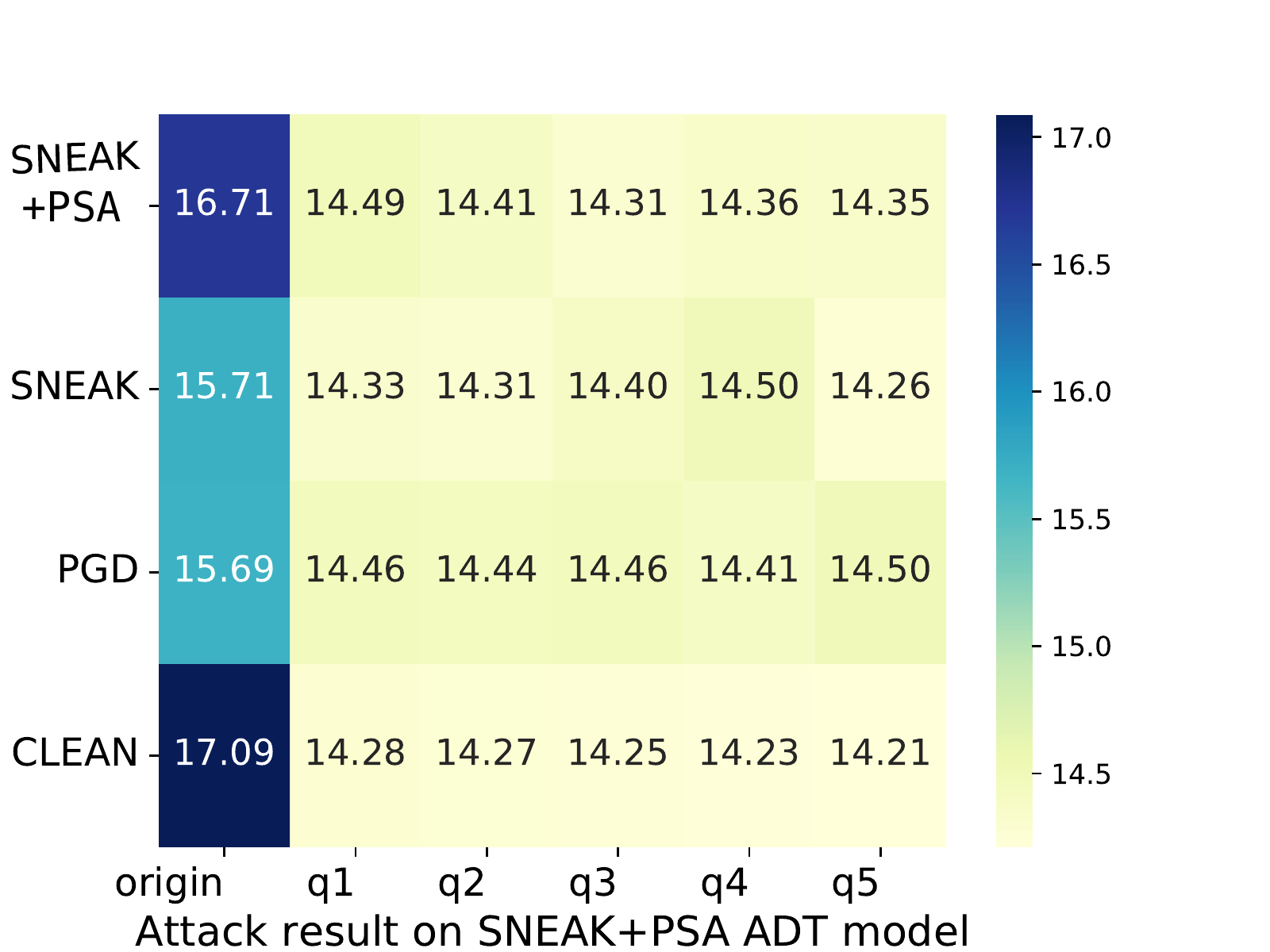}
    \end{minipage}
    \caption{Attack Result on models with Adversarial Training.}
    \label{fig defence}
       \vspace{-2em}
\end{figure}

In this part, we mainly compare SNEAK with PSA attack and SNEAK,  where varing settings for PSA is considered. Figure  \ref{fig:Qualitative} and \ref{fig:onecol} demonstrate a noticeable negative impact on attacking performance of SNEAK with PSA compared with the pure SNEAK attack. As can be seen, after pruning the $L2 \ norm$ of $\delta$ drops, which indicates that the $\delta$ generated with PSA is stealthier than it without. To tackle the drop on attack performance, we tried to increase either the clipping bound $B$ or $pruning \ size$ to enhance the attack capability. Table \ref{Table bound} indicates the the perturbation has larger impact on the model than a larger $B$, and the mIoU on test dataset drops steadily as $B$ grows, which eventually approaches the attack performance of the pure SNEAK attack.  Similar with the result of a larger bound, the model is more likely to be fooled as the $pruning \ size$ increases, as shown in Table \ref{Table size}.  This result is consistent with PSA's definition; as the $pruning \ size$ goes up, the PSA method's effect on SNEAK gradually declines, and eventually becomes identical to pure SNEAK attack when no clips are pruned.

\subsection{Defense with Adversarial Training}

Here, we demonstrate that adversarial training can mitigate the proposed SNEAK and SNEAK+PSA attacks. As shown in Figure \ref{fig defence}, it turns out that adversarial training only with PGD generated noise is limited to defend against PGD attack and still remain vulnerable to synonymous substitution attack. Meanwhile, adversarial training only with synonymous substitution attack is more effective in defense compared to the former one. However, the defense ability on the original query sentence is noticeably lower compared with the previous method by around 3 percent. The adversarial training only with perturbation generated by SNEAK provides more robust model, which shows greater defense performance on both the original and synonymous queries. Such defense capability improvement is mainly due to its utilization of both knowledge from video and query attack. This SNEAK adversarial training manages to beat other approaches in most of the situations. As for applying the SNEAK with PSA attack in adversarial training, since the defender can hardly predict on which clip the attacker tackle with noise and hence misleading the model, the overall defense  ability is weaker, which leads to  by $1 \% \sim2 \%$ accuracy drop compared with the one with SNEAK.

\section{Conclusion}
\label{sec.conc}

In this paper, we have inspected the vulnerability of NLVL model by proposing a new adversarial attack called SNEAK attack to deceive the target NLVL model with promising results. SNEAK is capable to utilize information from both video and language inputs, combined with the gradient splitting optimization to generate video perturbation. SNEAK has been proved more valid compared to single-modal attacks like video PGD attack or synonymous sentence substitution attack. To improve the stealthiness of video adversarial examples, we have further proposed a PSA attack method, which injects adversarial perturbations on partial frames with the highest impact on the localization prediction results. When combining SNEAK and PSA, it achieved comparable attack capability with SNEAK alone. We also study the complimentary aspect of the defense by proposing adversarial training for the SNEAK attack. A series of experiments conducted on the TACoS and Charades datasets proves the effectiveness of the new attacks and defense.

{\small
\bibliographystyle{ieee_fullname}
\bibliography{egbib}
}

\end{document}